%% file: root.tex
\documentclass[letterpaper, 10 pt, conference]{ieeeconf}  % Comment this line out if you need a4paper

\IEEEoverridecommandlockouts                              % This command is only needed if 
\usepackage{cite}
\usepackage{amsmath,amssymb,amsfonts}

\usepackage{graphicx}
\usepackage{textcomp}
\usepackage{xcolor}
\usepackage{subcaption}
\usepackage{algorithmicx}
\usepackage{algorithm}
\usepackage{algpseudocode}
\usepackage{float}

\title{\LARGE \bf
DNFOMP: Dynamic Neural Field Optimal Motion Planner for Navigation of Autonomous Robots in Cluttered Environment
}

\author{Maksim Katerishich, Mikhail Kurenkov, Sausar Karaf, Artem Nenashev, and Dzmitry Tsetserukou
\thanks{The authors are with the Intelligent Space Robotics Laboratory, Skoltech, Bolshoy Boulevard 30, bld. 1, 121205, Moscow, Russia }
\thanks{Email: \{Maksim.Katerishich, Mikhail.Kurenkov, Sausar.Karaf, Artem.Nenashev, D.Tsetserukou\}@skoltech.ru
}
}

\begin{document}

\maketitle
\thispagestyle{empty}
\pagestyle{empty}

\maketitle

\input{chapters/0_abstract.tex}

\input{chapters/1_introduction}
\input{chapters/2_related_work}

\input{chapters/3_methodology}
\input{chapters/4_experiments_and_results}
\input{chapters/5_conclusions_and_future_work}

\bibliographystyle{ieeetr}
\bibliography{bibliography.bib}

\end{document}

%% file: chapters/0_abstract.tex
\begin{abstract}
Motion planning in dynamically changing environments is one of the most complex challenges in autonomous driving. Safety is a crucial requirement, along with driving comfort and speed limits. While classical sampling-based, lattice-based, and optimization-based planning methods can generate smooth and short paths, they often do not consider the dynamics of the environment. Some techniques do consider it, but they rely on updating the environment on-the-go rather than explicitly accounting for the dynamics, which is not suitable for self-driving. To address this, we propose a novel method based on the Neural Field Optimal Motion Planner (NFOMP), which outperforms state-of-the-art approaches in terms of normalized curvature and the number of cusps. Our approach embeds previously known moving obstacles into the neural field collision model to account for the dynamics of the environment. We also introduce time profiling of the trajectory and non-linear velocity constraints by adding Lagrange multipliers to the trajectory loss function. We applied our method to solve the optimal motion planning problem in an urban environment using the BeamNG.tech driving simulator. An autonomous car drove the generated trajectories in three city scenarios while sharing the road with the obstacle vehicle. Our evaluation shows that the maximum acceleration the passenger can experience instantly is -7.5 m/s\textsuperscript{2} and that 89.6\% of the driving time is devoted to normal driving with accelerations below 3.5 m/s\textsuperscript{2}. The driving style is characterized by 46.0\% and 31.4\% of the driving time being devoted to the light rail transit style and the moderate driving style, respectively.

\end{abstract}

% \begin{IEEEkeywords}
% motion planning, neural network, self-driving, autonomous robot
% \end{IEEEkeywords}

%% file: chapters/1_introduction.tex
\section{Introduction}

\begin{figure}[t]
\centering
\includegraphics[width=0.42\textwidth]{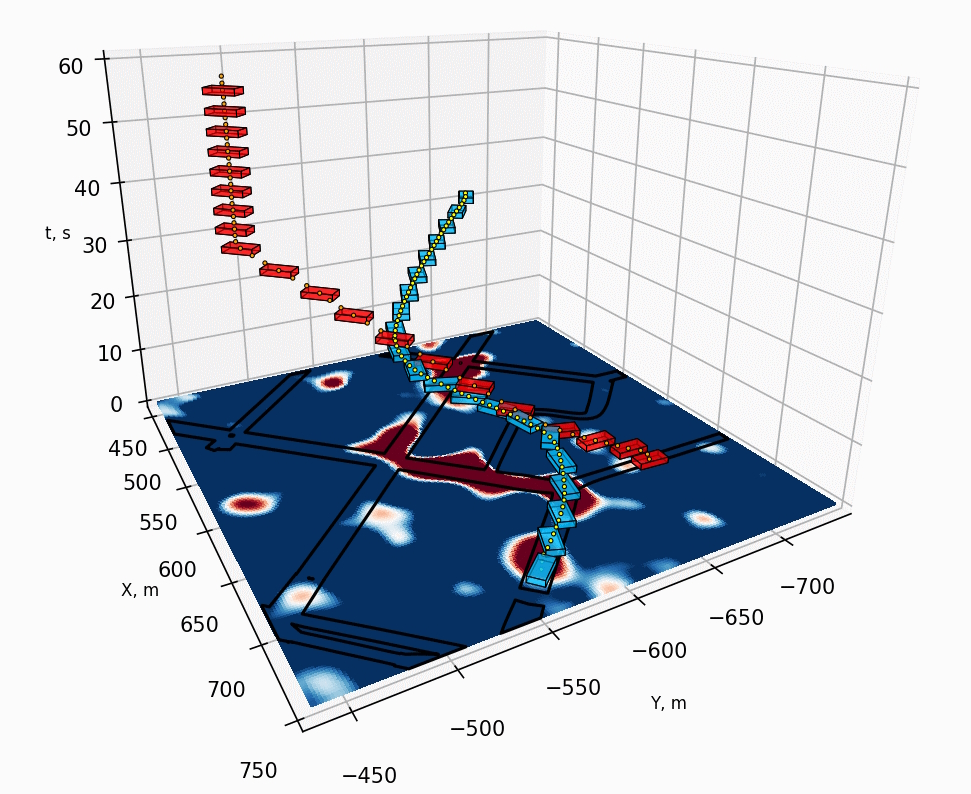}
\caption{DNFOMP graphical representation. Dark red is the low obstacle neural field value. Dark blue is the high obstacle neural field value. Blue and red bounding boxes represent sequential positions of the Ego vehicle and obstacle vehicle in time. Black lines are road edges.}
\label{fig:dnfomp_pic}
\vspace{-1.5em}
\end{figure}

Autonomous robotics is experiencing rapid growth, with major companies announcing plans to sell driving automation technologies. % \cite{8715479}. Despite some concerns, autonomous cars are highly favored due to their potential to provide safer, faster, more accessible, comfortable, and environmentally friendly transportation. 
However, ensuring automotive safety remains a key challenge for the development and commercialization of autonomous vehicles. Motion planning is crucial for self-driving cars to generate a safe and efficient path, but it is challenging due to the car's complex shape, non-holonomic constraints, and changing environments \cite{safety1}. 
%As such, motion planning is one of the core technologies for autonomous vehicles \cite{9266032}.

% Motion planning is a critical aspect of autonomous self-driving cars, involving the generation of a feasible and safe trajectory in a spatiotemporal space. The trajectory must be collision-free and optimized for parameters such as smoothness, length, comfort i.e. low accelerations, and travel time. However, modern motion planners face challenges in providing sufficient trajectory parameters while taking into account the changing environment. The task of motion planning for self-driving cars is complicated due to several factors. Firstly, the shape of the car is complex, requiring the calculation of the car's position as an element of SE(2) space instead of a simpler two-dimensional Euclidean space. Secondly, the motion patterns of autonomous vehicles must satisfy non-holonomic constraints. Therefore, the development of an algorithm for optimal motion planning in a cluttered environment for autonomous self-driving cars is crucial. This algorithm will enable the generation of a sequence of collision-free car positions that connect the start and goal points along a path that the car can traverse.

The recently presented NFOMP: Neural Field Optimal Motion Planner \cite{9851532} successfully overcomes most of the problems of sample-based, lattice-based, and optimization-based planners. It is significantly superior to the most state-of-the-art approaches, outperforming them by 25\% on normalized curvature and by
75\% on the number of cusps, but it was presented only in a static environment. To improve the NFOMP, we modified the trajectory optimizer to generate a time-profiled trajectory and incorporated known obstacle dynamics into the neural field collision model. A visualization of the planner is shown in Fig. \ref{fig:dnfomp_pic}. We demonstrate that the resulting trajectory has sufficient driving comfort parameters and has no collisions by launching an autonomous car to ride along several trajectories in the presence of other vehicles on the road in the BeamNG.tech simulator \cite{BeamNGTechnicalPaper21}. We collected data from the simulator's on-board accelerometer and showed that accelerations that affect a ``virtual passenger'' are admissible. 

\textbf{Thus, the main contributions of this work are:}
\begin{itemize}
    \item a neural field representation of collisions, that accounts for the moving obstacles;
    \item a trajectory optimization method for a self-driving car with velocity and steering constraints;
    \item an optimization-based algorithm for optimal motion planning of planar non-circular non-holonomic robots.
\end{itemize}

%% file: chapters/2_related_work.tex
\section{Related Work}

\begin{figure}
\vspace{0.75em}
\centering
  \hspace*{\fill}
  \begin{subfigure}{0.13\textwidth}
    \includegraphics[width=\linewidth]{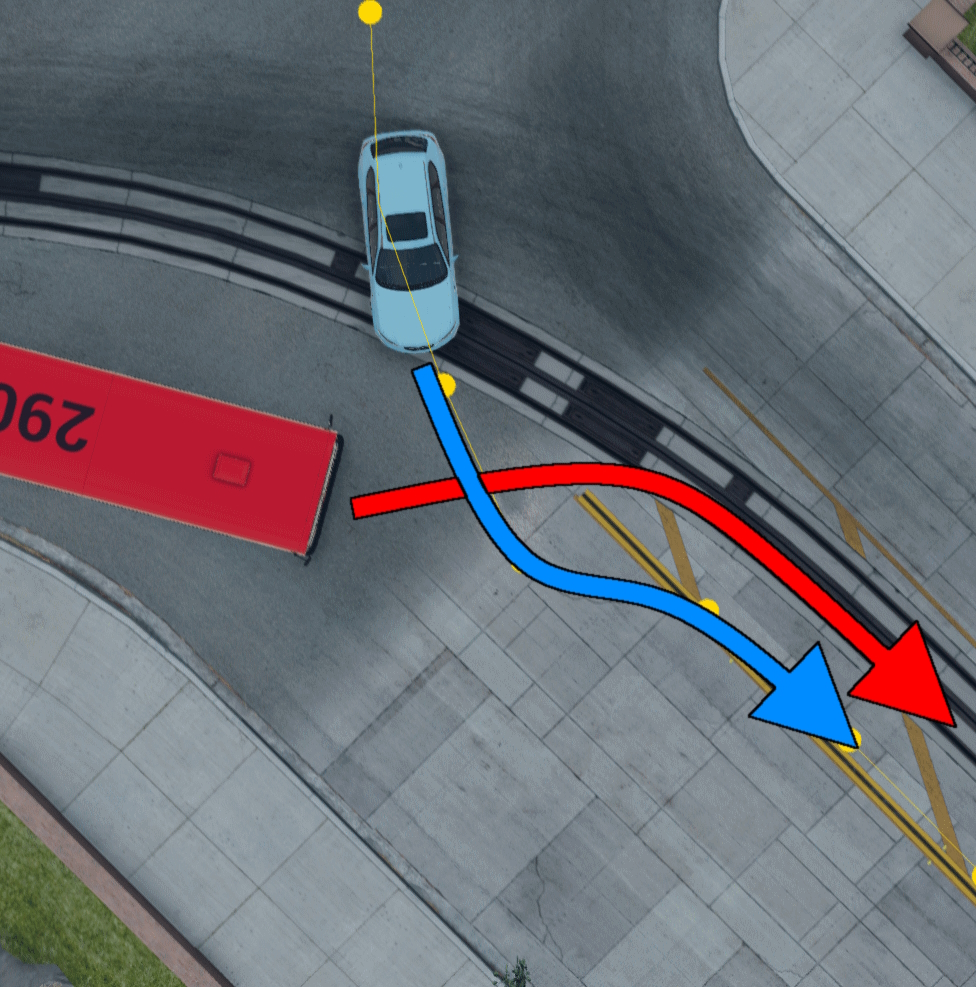}
    \caption{} \label{fig:continuous_1a}
  \end{subfigure}
  \hspace*{\fill}
  \begin{subfigure}{0.13\textwidth}
    \includegraphics[width=\linewidth]{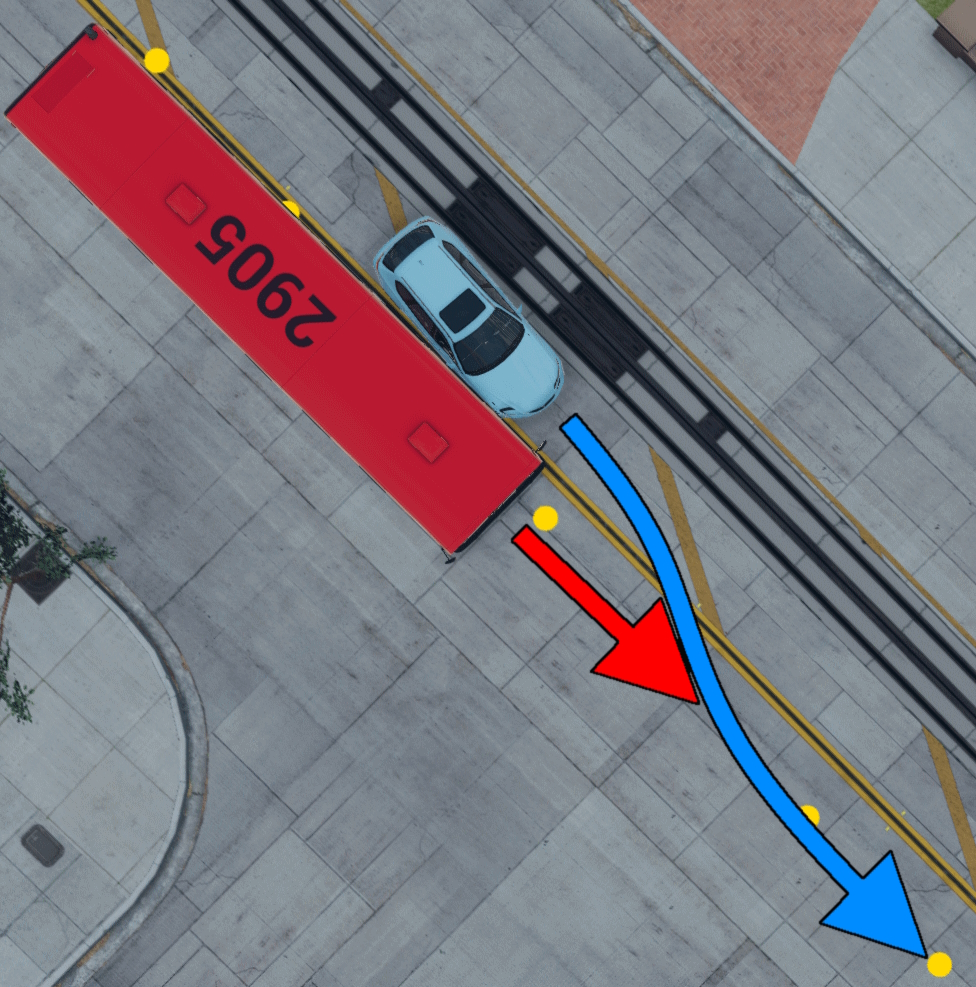}
    \caption{} \label{fig:continuous_1b}
  \end{subfigure}
  \hspace*{\fill}
  \\
    \hspace*{\fill}
    \begin{subfigure}{0.13\textwidth}
    \includegraphics[width=\linewidth]{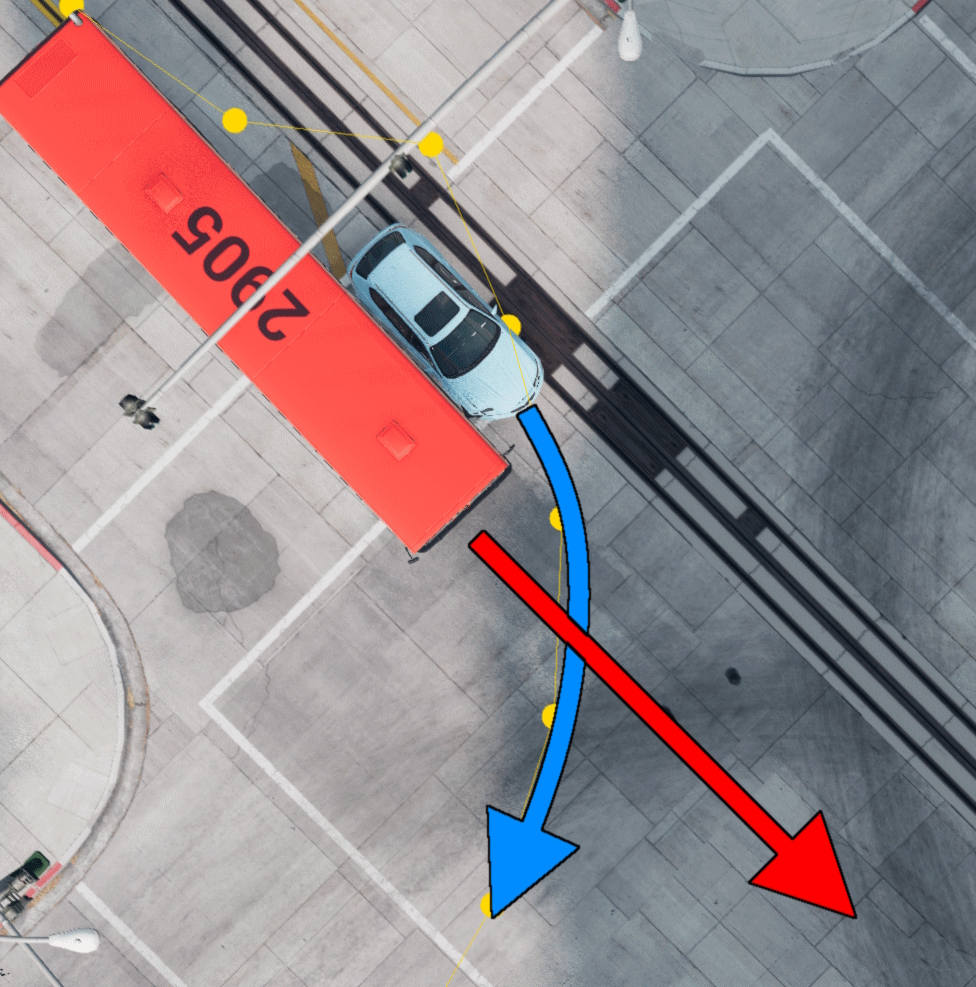}
    \caption{} \label{fig:continuous_1c}
  \end{subfigure}
  \hspace*{\fill}
    \begin{subfigure}{0.13\textwidth}
    \includegraphics[width=\linewidth]{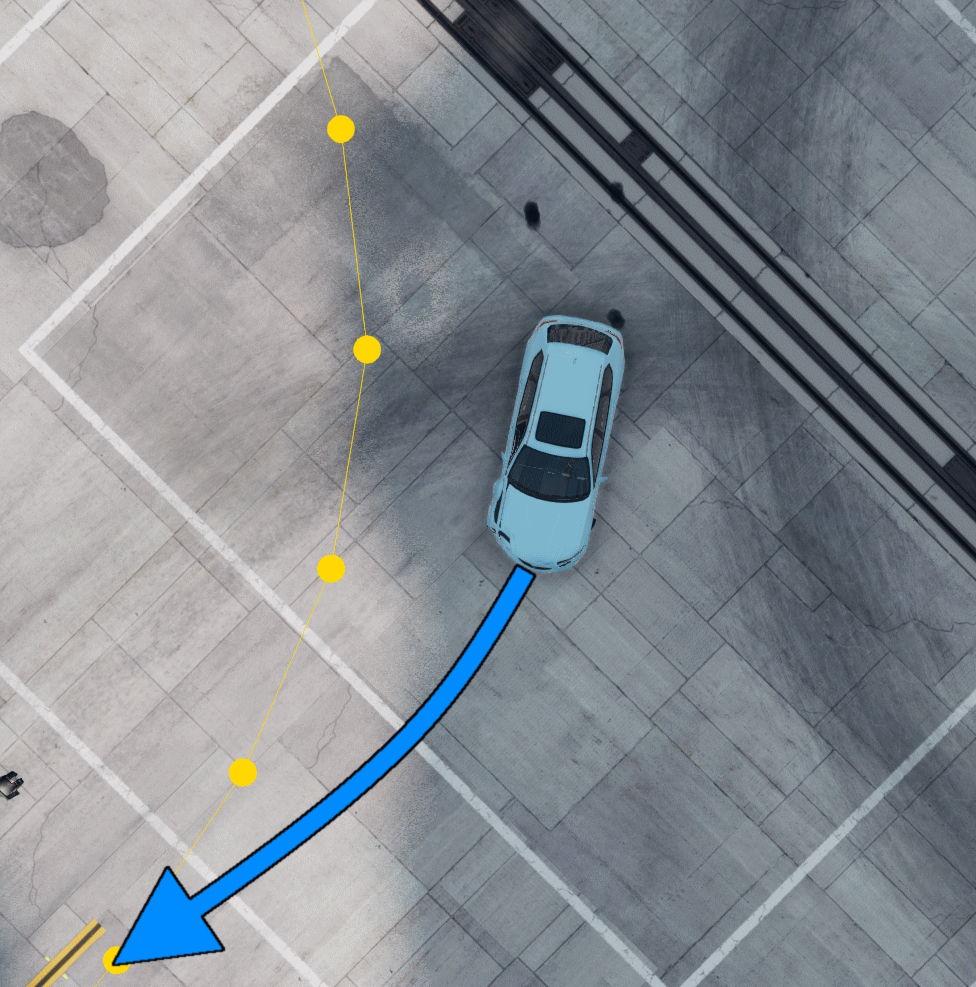}
    \caption{} \label{fig:continuous_1d}
  \end{subfigure}
  \hspace*{\fill}
\caption{Limitations of the continuous replanning methodology. The blue vehicle is the Ego vehicle, the red vehicle is the Obstacle vehicle, and the yellow line shows the planned trajectory. The blue and Red arrows show the direction of driving of the Ego and Obstacle vehicles, respectively.}
\label{fig:fig_continuous}
\vspace{-1.5em}
\end{figure}

An exact combinatorial solution for motion planning is impossible as it is proven to be PSPACE-hard \cite{lavalle_2006}. The only way to solve the motion planning problem is to find inexact solutions. The prior work in this area can be classified according to the approach used to find an initial path or optimize the prior trajectory. 
% \subsection{Random Sampling-Based Planning}

There are existing random sampling-based approaches to the problem of highway planning based on sampling in spatio-temporal space, such as ST-RRT\textsuperscript{*} \cite{9811814} and SCATE \cite{9811875}. These planners can quickly achieve a feasible path; however, they do not provide smooth and short trajectories, and further processing of such a trajectory is required \cite{9082624}. Moreover, adding more dimensions to the planning space generates an additional problem of computational complexity that should be conditioned and solved in special ways \cite{980223}. 
% \subsection{Lattice-Based Planning}

Another approach to this is to have a set of spline trajectories \cite{TALAMINO201993} or to build the trajectory using a conformal spatio-temporal lattice \cite{980223}. The benefit of this approach is that it ultimately produces smooth and plain trajectories, as they are essentially parametrized curves. However, such trajectories represent a predefined set, and the universal applicability of the approach is the main drawback \cite{KATRAKAZAS2015416}. 
% \subsection{Optimization-Based Planning}

Using neural networks is considered for the motion planning problem. The efficiency of such approaches is reached by combining them with other algorithms and using these networks as advisors for constrained problems, such as the selection of spline anchor points \cite{9812313} or recommending a local goal for an MPC \cite{9385847}. The main drawback of such an approach is still the requirement of supervised learning, which raises the question of validation under varying conditions. As one of the most outstanding examples, an optimization-based planner was used back in 2005 in Alice for the DARPA Grand Challenge \cite{darpa2005alice}. The commonly used approaches rely on artificial potential fields or precomputed signed distance functions \cite{ZIPS20161}, which are then improved by various algorithms to overcome known shortcomings. Also, they do not often account for dynamic obstacles \cite{lima2018optimizationbased}.

%% file: chapters/3_methodology.tex
\section{Methodology}

This section describes the proposed dynamic neural field optimal motion planner (DNFOMP). First, we describe the limitations of the continuous replanning methodology. Subsequently, we cover the proposed novel dynamic obstacle neural field model. Then, we discuss the added Lagrange multipliers and loss functions for trajectory optimization.

\subsection{Continuous replanning}

At first sight, the problem of a dynamic environment can be solved in a simpler way. If the positions of obstacles are updated in each planning step, then the movement of obstacles can be considered. Collision model loss is then an output of the continuously updated neural field with the BCE logit loss function as follows:

\begin{equation}\label{eq:eq01}
\resizebox{0.9\columnwidth}{!}{$
L_{onf}^{} = \frac{1}{N}\sum_{i=0}^{N}(BCElogit(F_{\Theta}^{j}(x_i, y_i, \theta_{i}), o_i), 
j = 0,1,2,...M
$}
\end{equation}

Here, we represent the SE (2) trajectory positions with $x$ and $y$ coordinates in the absolute frame, and $\theta$ heading angle; $F_{\Theta}^{j}(x_i, y_i, \theta_{i})$ is the neural network parametrized with $\Theta$ parameters, $N$ is the number of positions in the trajectory, $M$ is the horizon of the planning task, and $o_i$ is the ground truth collision measured by the collision function.

Such a straightforward approach does not perform well in a basic driving scenario. As shown in Fig. \ref{fig:fig_continuous}, the blue Ego vehicle is driving down the street and ready to turn right. The red obstacle vehicle drives in the same direction at the same speed. The relative positions of the vehicles are constant, and the planned trajectory can always be found ahead of the vehicles. The Ego vehicle is not able to turn unless the obstacle is far enough away and the optimizer is able to build another path. In certain cases, this can lead to aggressive braking, extensive steering, or a collision.

\subsection{Dynamic Neural Field Obstacle Model}

\begin{figure*}[t]
\vspace{0.75em}
\centering
\includegraphics[width=0.8\textwidth]{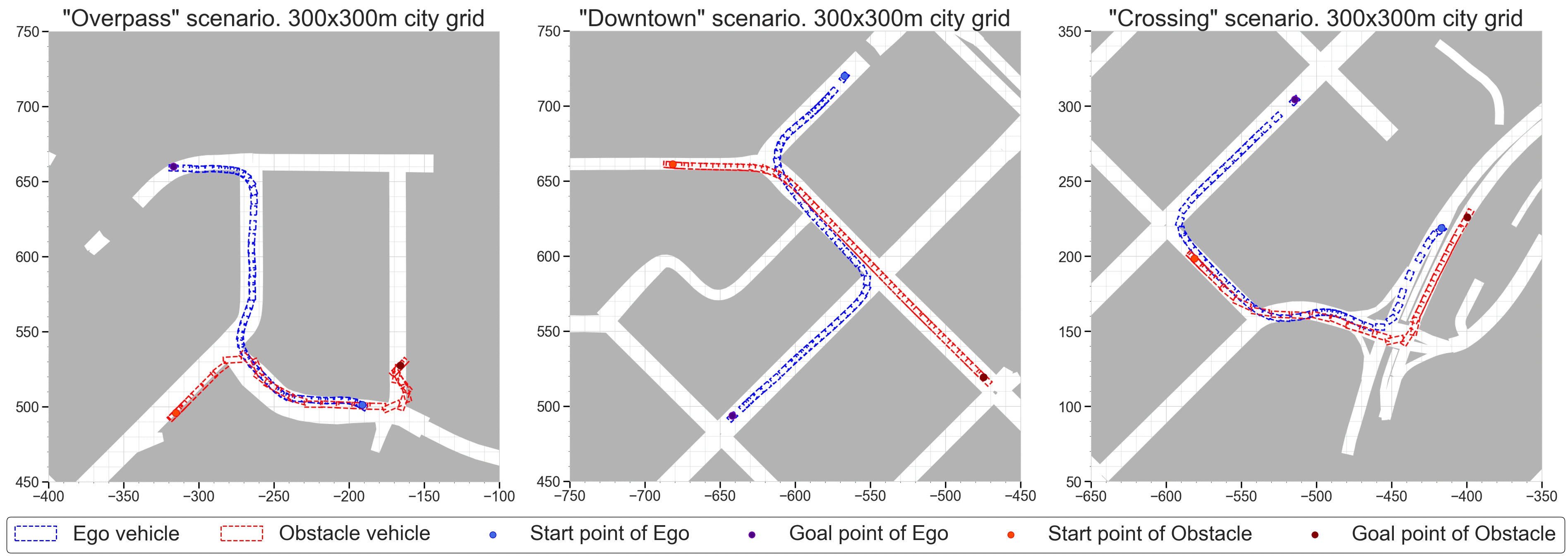}
\caption{Trajectories generated by DNFOMP in the test scenarios. Blue and red bounding boxes represent the sequential positions of the Ego and the obstacle vehicles, respectively. Gray and white are restricted and drivable zones, respectively.}
\label{fig:fig1}
%\vspace{-1.5em}
\end{figure*} 

We enable the collision model to consider not only the instant picture of the environment at a time step, but also the time-spanned collisions by extending the planning space. We introduce a new state variable, $t$, which stands for time. This makes all obstacles time-dependent. In this work, we set the trajectories of the obstacles based on prior knowledge. The adaptive nature of the neural field allows it to handle uncertainties; hence, the ground truth of the trajectory can be replaced with the estimation of one without any modification of the method but with hyperparameter tuning. We assume such prior knowledge of the obstacle trajectory to be sufficient for the method's validation. Thus, collision model loss is an output of the dynamic neural field with the BCE logit loss function as follows:

\begin{equation}\label{eq:eq10}
L_{dnf}^{} = \frac{1}{N}\sum_{i=0}^{N}(BCElogit(F_{\Theta}(x_i, y_i, \theta_{i}, t_i), o_i)
\end{equation}

Here, we represent trajectory positions with $x$, $y$ coordinates in the absolute frame, $\theta$ heading angle, and time $t$, which means that we may or may not have a collision at the same SE(2) point in the planning space, depending on the time when collision is checked. This neural network consists of three hidden layers with 128 neurons, each with a ReLU activation function. 

Thus, we introduce the ambiguity of the planning space. The trajectory optimizer can choose either to change the path itself in the spatial domain or modify the velocity profile to make the car brake or accelerate.

\subsection{Trajectory optimization in a dynamic environment}
Planning space is no longer planar but volumetric, where the third axis is time. To handle this ambiguity and consider that collision can be avoided in the spatial domain or in the temporal domain, we propose to add two more terms for trajectory loss building in addition to the original paper's trajectory loss terms. 
The first term is the velocity constraint. The goal of this constraint is to maintain the desired speed while driving. We use Lagrange multipliers that are multiplied by constraint deltas that should be zero and maximized during optimization. We used velocity constraint deltas $\delta_{i}$ from the equation:

\begin{equation} 
\resizebox{0.9\columnwidth}{!}{$
\delta_{i} = v_{e}(t_{i+1} - t_{i}) - \sqrt{(x_{i+1} - x_i)^2 + (y_{i+1}- y_i)^2 + (\theta_{i+1} - \theta_{i})^2}
\label{eq:eq20}
$}
\end{equation}

Here, $v_{e}$ is the desired velocity of the Ego vehicle. Thus, the velocity constraint loss $L_{vel}$ is given as follows:

\begin{equation}
L_{vel}^{} = \frac{1}{N}\sum_{i=0}^{N}(\delta_{i}^{2} - \lambda_{i} \delta_{i})
\label{eq:eq30}
\end{equation}

where $\lambda_{i}$ are the Lagrange multipliers.

The second term is the time regularization loss. The goal of this term is to penalize the uneven spacing of time intervals in the trajectory. This is aimed at minimizing longitudinal acceleration during driving. The time regularization loss $L_{time}$ is given as follows:

\begin{equation}
L_{time}^{} = \frac{1}{N}\sum_{i=0}^{N}(t_{i+1}-t_{i})^{2}
\label{eq:eq40}
\end{equation}

\begin{table}[t]
    \centering
    \caption{Numerical values of loss function weights}

        \resizebox{0.46\textwidth}{!}{
\begin{tabular}{p{7cm}p{1cm}}
    \textbf{Term} & \textbf{Value} \\
        Distance loss weight & $5 \cdot 10^{1}$ \\
        Collision loss weight & $5 \cdot 10^{4}$ \\
        Non-holonomic constraints loss weight& $5 \cdot 10^{1}$ \\
        Maximum velocity loss weight& $1 \cdot 10^{2}$ \\
        Time regularization loss weight& $1 \cdot 10^{2}$ \\
\end{tabular}
    \label{table1}
}
%\vspace{-3.0em}
\end{table}

Additionally, we include a modified trajectory reparametrization algorithm. In the original work, reparametrization is about generating a set of interpolated positions between successive trajectory points, to achieve a uniform spacing of positions. We first need to align time intervals and then sample interpolated positions. The newly distributed time values $t^{new}$ are:

\begin{equation}
t_{i}^{new} = i\frac{t_N}{N-1}, \quad i = 0,1,2,...N
\label{eq:eq50}
\end{equation}

Then, we calculate the collision loss for trajectory optimization as follows:

\begin{equation}
L_{col}^{} = \frac{1}{N-1}\sum_{i=0}^{N-1}(softplus(F_{\Theta}(\widetilde{x_i}, \widetilde{y_i}, \widetilde{\theta_{i}}, t_i^{new})))
\label{eq:eq60}
\end{equation}

where $(\widetilde{x_i}, \widetilde{y_i}, \widetilde{\theta_{i}})$ is the interpolated position randomly chosen between $(x_i, y_i, \theta_{i})$ and $(x_{i+1}, y_{i+1}, \theta_{i+1})$.

\subsection{Final Loss for Trajectory Optimization}
Thus, the final loss function consists of four parts: the collision loss function $L_{col}$, which pushes a trajectory away from collisions; the original paper trajectory loss terms, which are non-holonomic constraints $L_{constr}$ that force the trajectory to be smooth and Laplacian regularization $L_{dist}$ that shortens the overall path; velocity constraints $L_{vel}$ that force the vehicle to maintain constant speed along the trajectory; and time regularization $L_{time}$ that forces the velocity profile to be as plain as possible. The constraint loss, as well as the maximum velocity loss, includes two terms: quadratic and linear with Lagrange multipliers. For collision loss, we use the softplus function on the output of the neural network. Thus, the resulting trajectory loss can be calculated from:
\begin{equation}
\begin{split}
L_{traj}^{} = &w_{dist}L_{dist} + w_{col}L_{col}+ w_{constr}L_{constr} \\
&+w_{vel}L_{vel} + w_{time}L_{time}
\end{split}
\label{eq:eq70}
\end{equation}
where $w$ is the weight that corresponds to each of the terms. The numerical values are provided in Table \ref{table1}.
In the following section, we conducted a sensitivity analysis of how the weight  ratio affects the trajectory.

\subsection{Details of optimization}
The optimization is based on the Adam optimizer \cite{kingma2017adam} for both the obstacle neural field model and the trajectory loss. For the neural field obstacle model, we use learning rate of $1 \cdot 10^{-1}$ and betas of $(0.9, 0.9)$. In contrast to the original work, we use cyclic learning rate scheduling for trajectory optimization, which switches the learning rate between $1 \cdot 10^{-2}$ and $1 \cdot 10^{-1}$, with betas of $(0.9, 0.9)$. This helps to traverse the local minima faster and reduce the planning time. Learning rates are correlated to each other for stable planning. The Lagrange’s multipliers are optimized with the learning rate of $1 \cdot 10^{-1}$.

Gradients of the loss function are preconditioned based on an inverse of pseudo Hessian, as in the article \cite{hessian}, that is the sum of the quadratic form matrix and the identity matrix with coefficients. For gradient preconditioning, we use the following equation:

\begin{equation}
\begin{split}
g_i = \left[ \begin{array} {ll} \eta(\alpha H + I)^{-1} \triangledown (L_{traj}^{P}) \\ \eta(\alpha H + I)^{-1} \triangledown (L_{traj}^{T}) \end{array} \right] \\
\end{split}
\label{eq:eq80}
\end{equation}

Here $\eta$ is the learning rate; $\alpha$ is the hyperparameter that equals $5$; $H$ is the Hessians of $L_{traj}^{*}$, where $*$ stands for two different components: a $P$ part that carries the spatial path of the trajectory, i.e., all SE(2) points of it; and the second part $T$ that carries the temporal part of the trajectory, i.e., time values. The fact that we precondition these gradients separately without making a higher-dimensional Hessian allows us to tune trajectory loss function weights independently. Thus, $\eta(\alpha H + I)^{-1}$ is calculated once at the beginning of optimization, as Hessian is independent of an optimized trajectory.

For the validation of our method, the PyTorch framework is used for automatic differentiation. The developed code is available online at GitLab\footnote{https://gitlab.com/kukaruka/pytorch-motion-planner}.

%% file: chapters/4_experiments_and_results.tex
\section{Experiments and Results}

\subsection{Implementation Details}
To evaluate the performance of the proposed DNFOMP planner, we used the BeamNG.tech car simulator. It is a scenario-based testing environment for self-driving applications that thoroughly simulates the physics of driving. It has an "AI Vehicle" agent that controls a vehicle and follows the given trajectory. We launched this agent to ride along several trajectories in the presence of other vehicle.

The path-following algorithm of the agent is inaccessible due to its implementation in BeamNG.tech. Besides, the agent is unaware of the full trajectory. It follows the trajectory on a short horizon, a few positions ahead of the current one. It is essential for the trajectory to be uniformly parameterized both in the spatial and temporal domains, in order to let the agent traverse it. In fact, trajectory and agent are not coupled at all. It fails and causes collisions due to a high discrepancy between planned and executed positions if bad trajectory parameters are provided. Thus, this makes this agent a reasonable validation tool.

The following metrics were obtained: The first group depicts properties of the path: computation time in seconds (s); path length in meters (m); cusps are the number of abrupt changes of robot direction; the maximum and normalized curvature (lower is better); and angle-over-length (AOL) is the smoothness of the generated path. The second group shows the properties of driving along the trajectory. They are the longitudinal and lateral accelerations, which are obtained via the simulated accelerometer placed on the driver's headrest, and the trajectory following error, which is the Euclidean distance between actual and planned positions in time.

Experiments were conducted in the presence of a single moving obstacle vehicle. In each scene, five experiments with the same start and goal points were launched. The test cases included three different scenarios, which are 300x300 m city grids; see Fig. \ref{fig:fig1}. In each of them, a single moving obstacle is launched. A trajectory was initialized with the A\textsuperscript{*} algorithm. The video of the simulation is available on YouTube\footnote{https://youtu.be/yaeI4JM2-OI}.

% \item Driving in the presence of a multiple obstacles that drive in different directions relative to the Ego vehicle. On this scene, 5 experiments with different start and goal points were launched.

% iterations         cusps          AOL    norm_curv    max_curv    path_len    time
% ---------------  -------  -----------  -----------  ----------  ----------  ------
% nfomp100cross          0  -0.0010578      10.8892    0.126195      321.767   21.1
% nfomp200cross          1   0.00179006     15.9412    0.272376      331.384   43
% dnfomp700cross         0  -0.00085556      8.15293   0.0581912     302.495   38.1

% nfomp100hotel          0  -0.00641641     13.5645    0.13554       360.124   23.06
% nfomp200hotel          0  -0.0100504      12.5875    0.0710806     366.395   44.8
% dnfomp700hotel         0  -0.00911093     10.1861    0.0702038     344.736   41.9

% nfomp100harbor         0   0.00111534     17.0348    0.217786      284.504   19.2
% nfomp200harbor         0   0.00162984     17.5725    0.435817      281.515   38.4
% dnfomp700harbor        0  -0.00210381     13.4085    0.105515      264.901   36.8

\begin{table}[t]
\vspace{0.75em}
    \centering
    \caption{Planning statistics for the test scenarios}
    \label{table3}
    \resizebox{0.485\textwidth}{!}{
    \begin{tabular}{llllllll}
        \textbf{Planner} &\textbf{Time, s} &\textbf{Path length, m} &\textbf{Cusps} &\textbf{Max curv.} &\textbf{Norm curv.} & \textbf{AOL}$\cdot$ 10\textsuperscript{-3} \\
        \multicolumn{7}{l}{\textbf{Scenario:} Downtown} \\
            NFOMP & 32.6 & 321.7 & 1 & 0.12 & 10.9 & 1.05 \\
            DNFOMP & 38.1 & 302.5 & 0 & 0.05 & 8.15 & 0.85 \\
        \multicolumn{7}{l}{\textbf{Scenario:} Crossing } \\
            NFOMP & 34.2 & 360.1 & 0 & 0.07 & 12.6 & 1.0 \\
            DNFOMP & 41.9 & 344.7 & 0 & 0.07 & 10.2 & 0.9 \\
        \multicolumn{7}{l}{\textbf{Scenario:} Overpass } \\
            NFOMP & 28.8 & 281.2 & 0 & 0.21 & 17.0 & 1.1 \\
            DNFOMP & 36.8 & 264.9 & 0 & 0.10 & 13.4 & 2.1 \\
    \end{tabular}}
%\vspace{-1.5em}
\end{table}

\begin{figure}
\captionsetup[subfigure]{justification=centering}
  \begin{subfigure}{0.2\textwidth}
    \includegraphics[width=\linewidth]{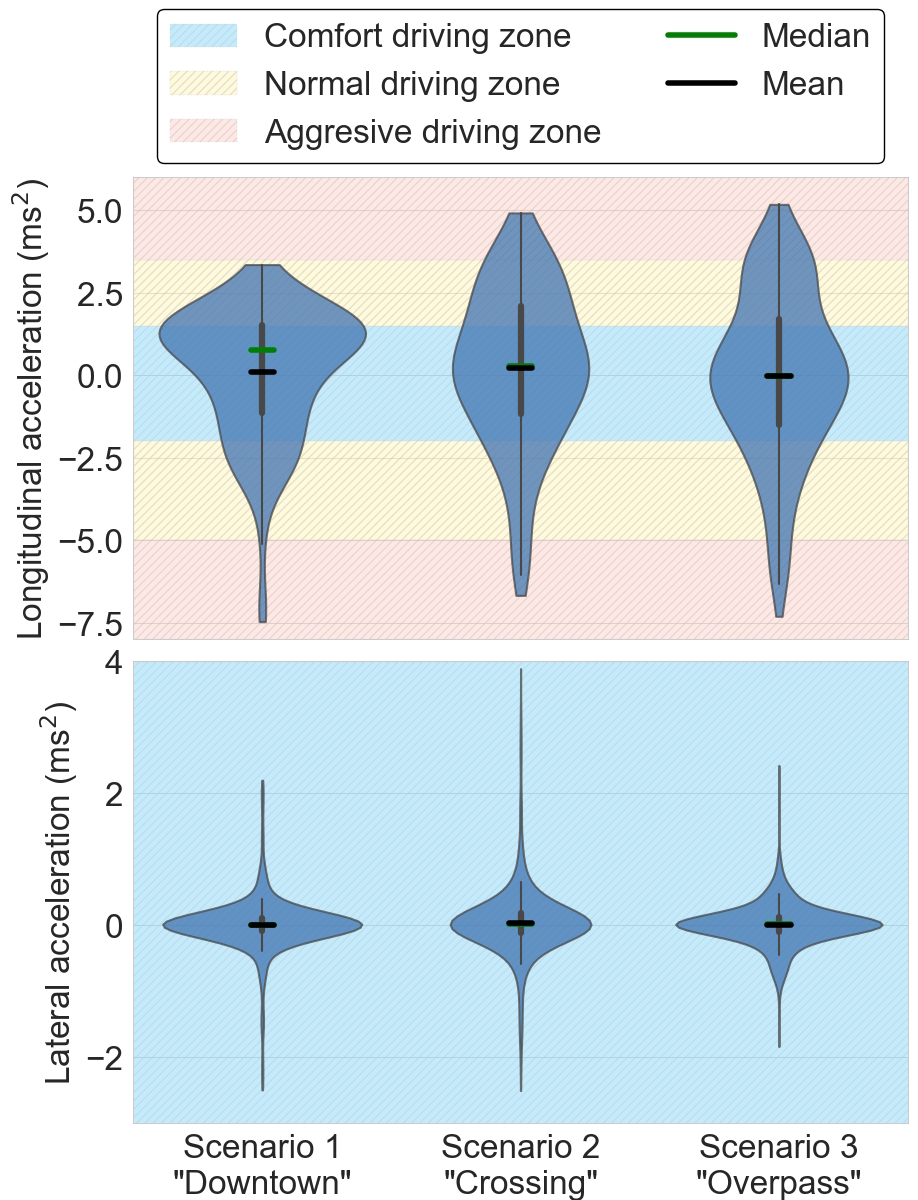}
    \caption{Driving comfort \\ assertion.} \label{fig:dynamics_1a}
  \end{subfigure}%
  \hspace*{\fill}   % maximize separation between the subfigures
  \begin{subfigure}{0.2\textwidth}
    \includegraphics[width=\linewidth]{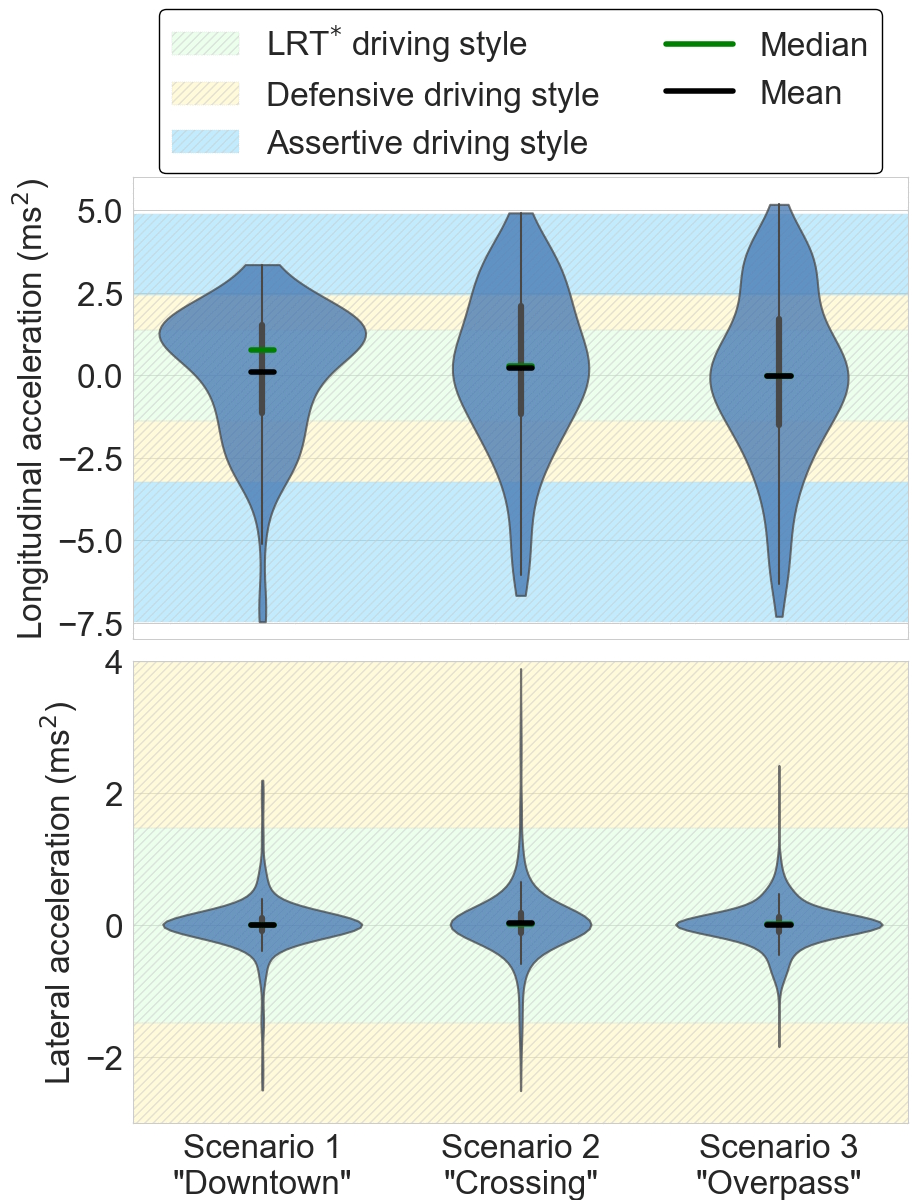}
    \caption{Driving style \\ assertion.} \label{fig:dynamics_1b}
  \end{subfigure}%
  \hspace*{\fill}   % maximizeseparation between the subfigures
    \caption{{Lateral and longitudinal accelerations of the Ego vehicle while driving the planned trajectories. \\ \textsuperscript{*} Light Rail Transport.}} \label{fig:dynamics}
%\vspace{-1.5em}
\end{figure}

\subsection{Path Properties Evaluation}
We compared the proposed DNFOMP and the original NFOMP in light of the metrics of the first group. For the NFOMP, we did not take any moving obstacles into account. Also, we did not include metrics of the second group, as they do not correspond to the same driving modes: without the presence of obstacles, the optimal spacing of the values is linear, while the presence of obstacles requires deforming the trajectory and using braking and acceleration. The results are presented in Table \ref{table3}. The proposed planner is able to keep similar trajectory metrics together with moving obstacle collision avoidance. Moreover, the method is dimension-agnostic, as the increased number of dimensions of the planning space did not affect planning time significantly.

\subsection{Dynamic Properties Evaluation}
This section presents the results of the experiments in light of the metrics of the second group. The desired speed is set to 40 kph.
The Ego vehicle car model is set to ``ETK-800'', which corresponds to a mid-range wagon that is capable of producing up to 1G overload (9.81 m/s\textsuperscript{2} acceleration magnitude) during driving with this setup. Full braking was allowed, and default throttle strength was enabled for the vehicle. For the evaluation, we used assigned zones with respect to accelerations.

The comfort, normal, and aggressive ranking of driving an autonomous shuttle bus is proposed in \cite{electronics8090943}. The authors opted for a driving style that would be accepted by a cautious passenger. According to this study, the interquartile range of longitudinal acceleration is within the normal driving zone in all test scenarios. The maximum acceleration reaches 5.0 m/s\textsuperscript{2}, and the maximum deceleration (braking) reaches -7.5 m/s\textsuperscript{2}, which are within acceptable zones. Based on longitudinal acceleration, on average 54.9\%, 34.7\%, and 10.3\% of driving time are devoted to comfort, normal, and aggressive driving, respectively. The evaluation is presented in Fig. \ref{fig:dynamics_1a}.

\begin{table}
\vspace{0.75em}
    \centering
    \caption{Relation between control stiffness of the agent and vehicle settings}

    \resizebox{0.42\textwidth}{!}{
    \begin{tabular}{lll}

        \textbf{Control stiffness} &\textbf{Brake force multiplier} &\textbf{Throttle strength multiplier} \\

            1.0 & 1.0 & 1.0 \\
            0.7 & 0.7 & 0.22 \\
            0.5 & 0.5 & 0.12 \\
            0.2 & 0.2 & 0.08 \\
            0.1 & 0.1 & 0.05 \\

    \end{tabular}
    }
        \label{table10}
\vspace{-1.5em}
\end{table}

Another categorization is proposed in \cite{accelerations2}. The authors assess the driving style of autonomous vehicles based on the feedback of human drivers who were passengers in the vehicle. Participants agreed with the defensive (moderate) style and found the light rail transit driving style too tedious and the assertive style too rough. Based on longitudinal acceleration, the driving style of the agent corresponds to the light rail transport style at 46.0\% of driving time, to the defensive and assertive styles at 31.4\% and 26.6\% of driving time, respectively. The evaluation is presented in Fig. \ref{fig:dynamics_1b}.

\begin{figure}[t]
\vspace{0.75em}
\centering
\includegraphics[width=0.4\textwidth]{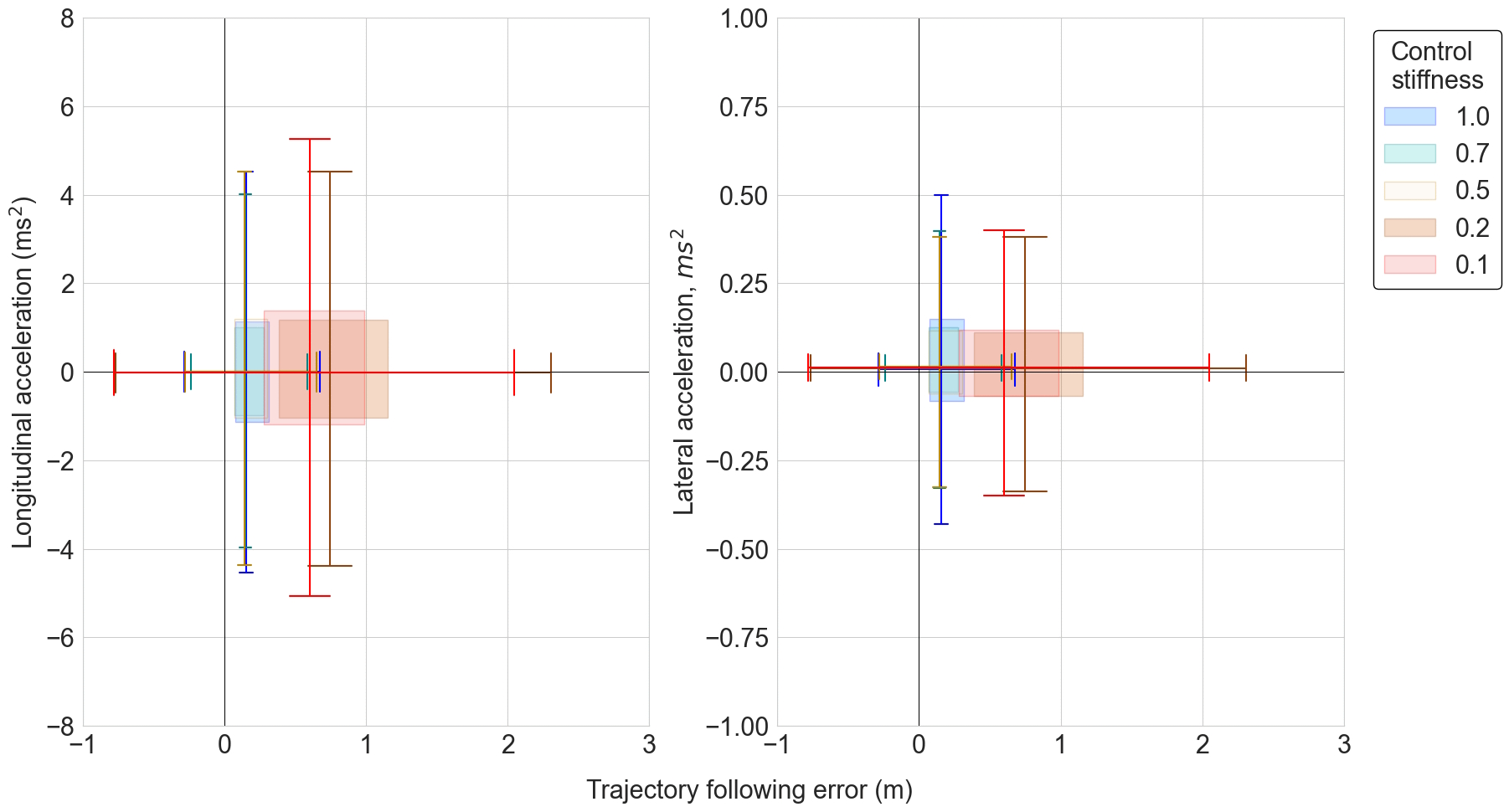}
\caption{Trajectory following error and corresponding accelerations of the Ego vehicle depending on control stiffness.}
\label{fig:boxplots2d}
%\vspace{-1.5em}
\end{figure}

We validated the quality of trajectory parametrization by introducing the ``control stiffness" parameter, which is an aggregate of explicitly limiting brake force and implicitly reducing throttle strength of the vehicle, as shown in Table \ref{table10}. With 1.0 control stiffness, the maximum trajectory following error is 0.67 m and the maximum longitudinal acceleration is 4.5 m/s\textsuperscript{2}. While reducing the stiffness down to 0.5, the acceleration remains above 4 m/s\textsuperscript{2} and the error is slightly increased. With the significant reduction in control stiffness down to 0.2, the error increases rapidly to 2.5 m. Therefore, the trajectory is parametrized in such a way that the agent executes it with the desired parameters while having the capability to drive more aggressively. Unless this capability is below the requirements of the planning algorithm, it succeeds in providing considerable route following, as shown in Fig. \ref{fig:boxplots2d}.

\begin{figure}[t]
\centering
\includegraphics[width=0.3\textwidth]{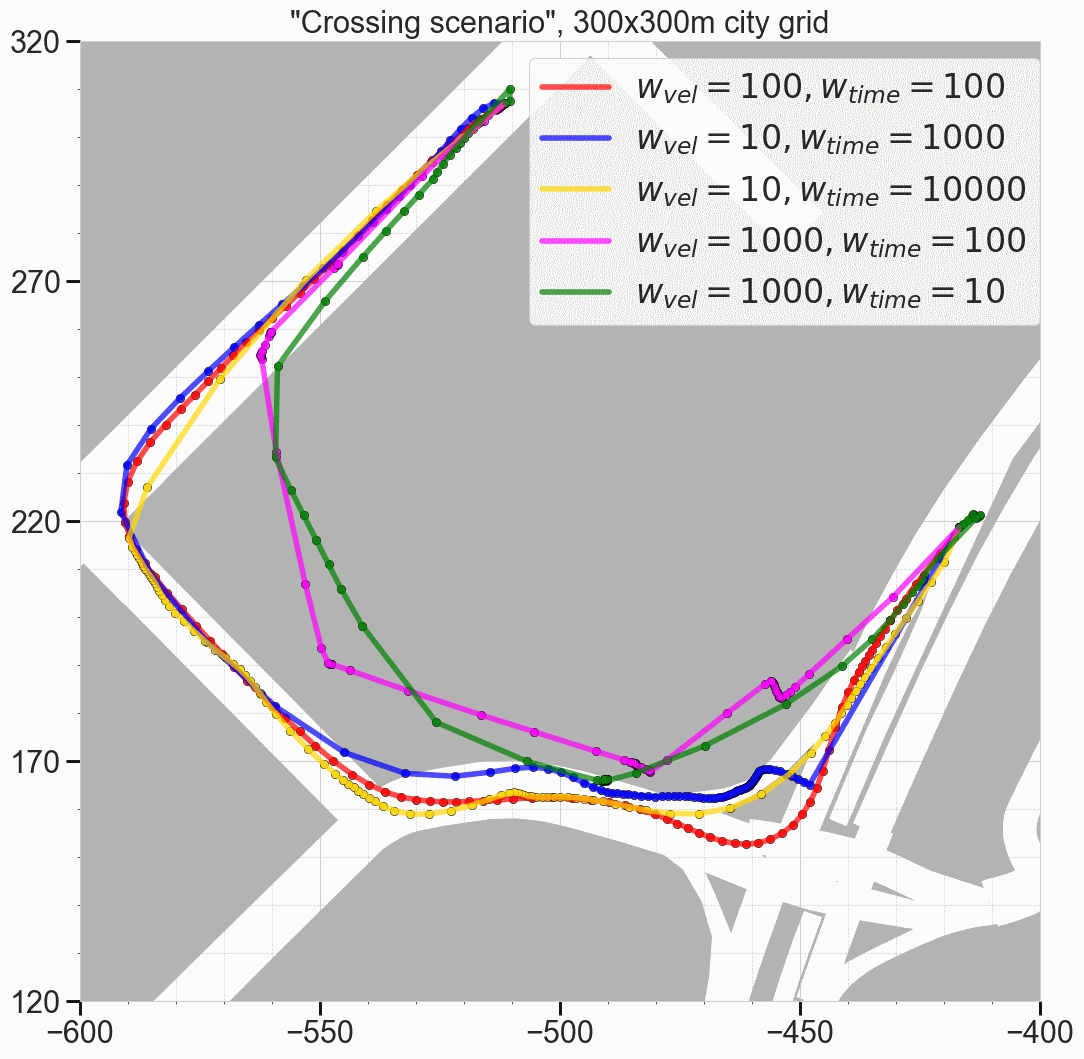}
\caption{Sensitivity analysis of different parameters.}
\label{fig:sensivity}
\vspace{-1.5em}
\end{figure}

\subsection{Sensitivity analysis}
We conducted a sensitivity analysis to show the influence of time regularization weight and velocity constraint weight on trajectory quality. Fig. \ref{fig:sensivity} shows that the ratio of these weights should be close to 1. If the time regularization weight dominates the velocity constraint weight, the trajectory becomes ill-parametrized in the space domain. The result is point clustering around high collision zones, where the velocity is lower than in the rest of the trajectory. If the velocity constraint weight dominates the time regularization weight, the trajectory becomes ill-parametrized in the time domain, and shortcutting can appear. The result is trajectory shrinking and point clustering around zones where time values are close.

%% file: chapters/5_conclusions_and_future_work.tex
\section{Conclusion and Future Work}
We have presented a novel algorithm to solve the optimal motion planning problem for a self-driving car in cluttered urban environments. This algorithm, due to embedding environment dynamics into the neural field collision model, makes it possible to plan a trajectory in spatio-temporal space and avoid moving obstacles. To validate the proposed method, an autonomous vehicle drove along the planned trajectories in the BeamNG.tech driving simulator. Conventional metrics such as trajectory curvature and the number of cusps were obtained. In these terms, the proposed approach is at the same level as the NFOMP baseline, and moving obstacles are considered additionally. According to the study of dynamic properties, the maximum acceleration that a passenger can experience instantly is -7.5 m/s\textsuperscript{2}, and 89.6\% of the driving time is devoted to normal driving with accelerations below 3.5 m/s\textsuperscript{2}; the assertion of driving style shows that 46.0\% and 31.4\% of the driving time are devoted to the light rail transit style and the moderate driving style, respectively.

In the future, we plan real-world experiments on various setups, for example, outdoor \cite{protasov2021cnn, karpyshev2021autonomous, karpyshev2022mucaslam} and indoor \cite{mikhailovskiy2021ultrabot, perminov2021ultrabot, petrovsky2020customer, okunevich2021deltacharger} mobile robots, UAVs \cite{kalinov2021impedance, kalinov2019high, kalinov2020warevision, kalinov2021warevr, yatskin2017principles}, as well as examine the applicability of the approach to more sophisticated systems, e.g., modular two-wheeled rovers \cite{petrovsky2022two}.

%The disadvantage of the proposed method is a relatively large computation time. However, the optimization of the collision model and trajectory takes less than 10\% of the total time. The rest of the time is devoted to collision checking that can be efficiently accelerated with a GPU parallel computing. Further improvement of the method may be comprising of a car dynamics and considering specific car kinematic constraints directly.